\pdfoutput=1

\documentclass[letterpaper, 10 pt, journal, twoside]{IEEEtran}

\IEEEoverridecommandlockouts                              

\newcommand{\RALreceived}{July 1, 2026}
\newcommand{\RALrevised}{September 28, 2026}
\newcommand{\RALaccepted}{November 4, 2026}
\newcommand{\RALeditor}{Sehoon Ha}
\newcommand{\RALvideoid}{U1tWZzHgv60}
\providecommand{\finalchange}[1]{#1}

\usepackage{graphics} 
\usepackage{epsfig} 
\usepackage{mathptmx} 
\usepackage{times} 
\usepackage{amsmath} 
\usepackage{amssymb}  
\usepackage{mathtools} 
\usepackage{xcolor,colortbl}
\usepackage{booktabs}
\usepackage{tabularx}
\usepackage{multirow}
\usepackage{algorithm}
\usepackage{algpseudocode}
\usepackage{url}
\usepackage{cite}
\usepackage{setspace}
\usepackage{needspace}
\usepackage{tikz}
\usepackage{bm}
\usepackage[hidelinks]{hyperref}
\usepackage{xurl}
\usetikzlibrary{arrows.meta, positioning, calc, decorations.pathreplacing, fit, backgrounds}

\usepackage[caption=false,font=footnotesize]{subfig}
\usepackage{stfloats} 

\makeatletter
\let\oldthebibliography\thebibliography
\renewcommand{\thebibliography}[1]{%
  \oldthebibliography{#1}%
  \setlength{\itemsep}{0pt}%
  \setlength{\parskip}{0pt}%
}
\makeatother

\title{Learned Controllers for Agile Quadrotors\\in Pursuit-Evasion Games}

\author{Alejandro S\'anchez Roncero, Yixi Cai, Olov Andersson and Petter \"Ogren
\thanks{Manuscript received: \RALreceived; Revised: \RALrevised; Accepted: \RALaccepted.}%
\thanks{This paper was recommended for publication by Editor \RALeditor{} upon evaluation of the Associate Editor and Reviewers' comments.
This work was partially supported by the Wallenberg AI, Autonomous Systems and Software Program (WASP) funded by the Knut and Alice Wallenberg Foundation.}%
\thanks{The authors are with the Department of Robotics, Perception and Learning, School of Electrical Engineering and Computer Science, KTH Royal Institute of Technology, SE-100 44 Stockholm, Sweden. Corresponding author: \href{mailto:alesr@kth.se}{\texttt{alesr@kth.se}}.}%
\thanks{Digital Object Identifier (DOI): see top of this page.}%
}

\definecolor{lavender}{rgb}{0.9, 0.9, 0.98}
\definecolor{red}{RGB}{255, 0, 0}
\definecolor{orange}{RGB}{252, 130, 62}
\definecolor{blue}{RGB}{0, 0,255}
\definecolor{darkgreen}{RGB}{0, 150,0}

\usepackage{eso-pic}
\AddToShipoutPictureFG*{\AtPageLowerLeft{\put(\LenToUnit{0.6in},\LenToUnit{0.22in}){\parbox[b]{\dimexpr\paperwidth-1.2in\relax}{\scriptsize\centering \copyright~2026 IEEE. Personal use of this material is permitted. Permission from IEEE must be obtained for all other uses, in any current or future media, including reprinting/republishing this material for advertising or promotional purposes, creating new collective works, for resale or redistribution to servers or lists, or reuse of any copyrighted component of this work in other works.}}}}
\begin{document}

\maketitle

\suppressfloats[t]
\begin{abstract}
We study one-versus-one (1v1) quadrotor pursuit--evasion, where a pursuer and an evader are trained with reinforcement learning (RL). Direct adversarial RL tends to be unstable because opponent policies change during training, causing non-stationarity, forgetting, and strategy cycling. We propose Asynchronous Multi-Stage Population-Based training, inspired by Policy-Space Response Oracles (PSRO). Proximal Policy Optimization (PPO) trains responses against previously learned opponents held fixed during each stage. The uniform variant, AMSPB-U, samples opponents equally, while AMSPB-P combines hard prioritized fictitious self-play (PFSP) with uniform exploration to emphasize currently difficult opponents. In cross-evaluations against learned opponents, both variants improve pursuit--evasion balance over latest-opponent training. AMSPB-P scores highest while retaining competence against earlier strategies. Across varied starting positions and after larger-arena retraining, both population variants retain higher average success across the two roles than latest-opponent training. Hardware experiments with two 10~cm, approximately 40~g Crazyflie brushless quadrotors demonstrate zero-shot sim-to-real transfer against baseline controllers and learned adversaries across 42 flights. Recorded flights include captures of evaders reaching robust peak speeds of 3.31~m/s and accelerations of 17.95~m/s\textsuperscript{2}, with pursuer flight speeds up to 4.87~m/s.
\end{abstract}


\section{INTRODUCTION} \label{sec:introduction}

\begin{figure}[!t]
    \centering
    \def\svgwidth{\columnwidth}
    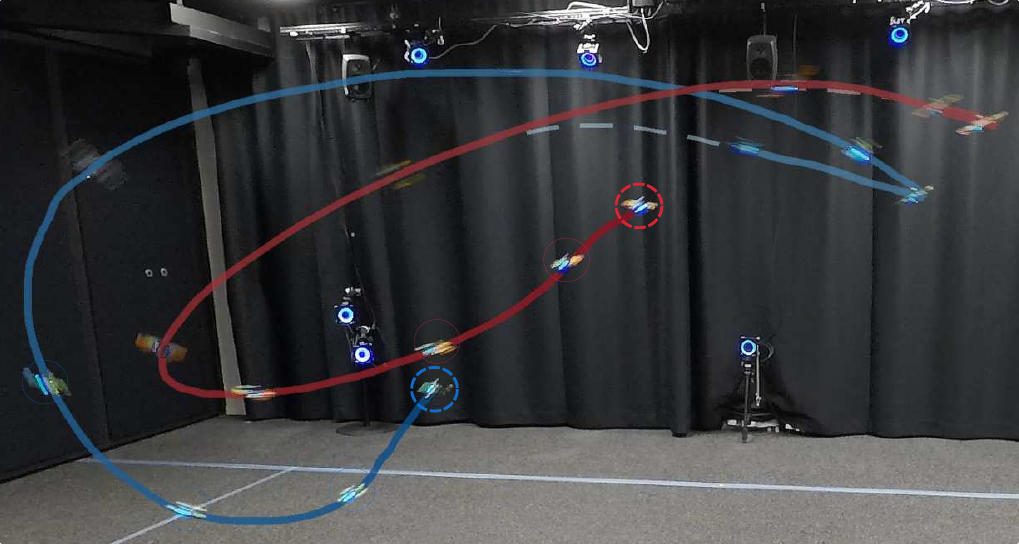
    \caption{
    Representative three-second hardware sequence with learned pursuit and evasion controllers. The pursuer (blue) is trying to catch the evader (red).
    }
    \label{fig:first_figure}
\end{figure}

The increasing proliferation of small unmanned aerial vehicles (UAVs) such as quadrotors has raised urgent safety and security concerns \cite{parkSurveyAntiDroneSystems2021}.
To mitigate these risks, we explore strategies where a friendly pursuer UAV intercepts an intruder (see Fig.~\ref{fig:first_figure}).
We focus on the strategy-learning problem between equal-capability vehicles, defining capture by proximity within a capture radius rather than testing a physical capture device.

In general, intercepting a maneuvering quadrotor is challenging: its strategy and dynamics are often unknown a priori, and quadrotors are highly maneuverable \cite{songReachingLimitAutonomous2023}.
Pursuit--evasion games (PEG) naturally model such interactions.
The central challenge is developing a pursuit strategy that generalizes across diverse adversarial behaviors.

Reinforcement learning (RL) offers a promising approach: one specifies the task through a reward function (capture for the pursuer, survival for the evader) and trains a neural network policy that maps state observations directly to body-rate and thrust commands \cite{kaufmann2022benchmark}.
This has proven effective in drone racing, where RL controllers have surpassed expert human pilots by better exploiting the quadrotor's dynamic envelope \cite{songReachingLimitAutonomous2023, hanoverAutonomousDroneRacing2024}.

A natural option is adversarial training, where both agents' RL policies compete during training, generalizing self-play to a two-player setting that has led to emergent behavior in other domains \cite{vinyals2019grandmaster}.
However, adversarial RL introduces challenges such as non-stationarity, catastrophic forgetting, strategy cycling, and policy collapse \cite{de2020independent, jaderberg2017population}.
\cite{xiaoLearningMultipursuitEvasion2024} proposed Asynchronous Multi-Stage Deep RL (AMSDRL), where agents train asynchronously across stages to alleviate non-stationarity.
However, since each stage trains against only the latest opponent, the approach remains prone to forgetting and cycling.

We propose \textit{Asynchronous Multi-Stage Population-Based training}, a Policy-Space Response Oracles (PSRO)-style approach for adversarial quadrotor RL~\cite{lanctot2017unified, mullerGeneralizedTrainingApproach2020}.
In our setting, each strategy is a neural quadrotor controller executed through body-rate and thrust commands in a randomized dynamics simulation.
At each generation, PPO trains against frozen opponent mixtures. Our uniform variant, AMSPB-U, samples opponents equally, while AMSPB-P prioritizes difficult opponents through hard prioritized fictitious self-play (PFSP)~\cite{vinyals2019grandmaster} mixed with uniform exploration.
Our contributions are:
\begin{itemize}
\item We adapt population-based response-oracle training to quadrotor pursuit--evasion, representing strategies as PPO-trained body-rate controllers.

\item We compare uniform and prioritized population training, showing improved retention of earlier skills and generalization across learned opponents and baseline controllers relative to training against only the latest opponent.

\item We validate the trained policies on two real $10$~cm Crazyflie brushless quadrotors without hardware fine-tuning, demonstrating zero-shot sim-to-real transfer against baseline, learned, and manually controlled opponents across $42$ flights.
\end{itemize}

\noindent\textit{Supplementary video:}
\href{https://youtu.be/\RALvideoid}{\texttt{youtu.be/\RALvideoid}}.

\Needspace{4\baselineskip}
\section{RELATED WORK} \label{sec:related_work}

Pursuit--evasion games have been studied from three main perspectives: model-based control, learning-based methods, and game theory.

\paragraph{Classical and model-based approaches}
Foundational differential-game work~\cite{isaacs1999differential} established optimal strategies for continuous-time pursuit--evasion under simplified dynamics.
Extensions include pursuit--evasion with obstacles \cite{fisac2015pursuit} and multi-pursuer settings \cite{fang2020cooperative}.
These methods provide guarantees under assumptions about dynamics and information, complementing approaches that learn responses through interaction.
In model-based control, \cite{ghotavadekarVariableTimeStepMPC2025} proposed variable time-step model predictive control (MPC) for intercepting dynamic targets, and \cite{pliskaSafeMidAirDrone2024} developed fast-response proportional navigation (FRPN) based on the current opponent state.
Sampling-based online planning combines target-motion prediction, learned trajectory timing and online path selection~\cite{ryou2025interception}. These approaches update interception plans as target observations change, without jointly training the target's evasion strategy.

\paragraph{Reinforcement learning for pursuit--evasion}
Multi-agent RL works \cite{duCooperativePursuitUnauthorized2021, kouzegharMultiTargetPursuitDecentralized2023} addressed cooperative pursuit but relied on simplified kinematics.
\cite{linDistributedPursuitEvasion2025} proposed curriculum learning based on increasing difficulty, and
\cite{chenOnlinePlanningMultiUAV2025} studied multi-UAV capture with real brushed Crazyflie 2.1 pursuers limited to $1.0$~m/s and a virtual non-learned evader.
\cite{xiaoLearningMultipursuitEvasion2024} proposed Asynchronous Multi-Stage Deep RL (AMSDRL), alternating training between pursuers and an evader while freezing the opponent at each stage.
AMSDRL is an important staged-training baseline, but trains each stage only against the latest opponent.
Several RL-based pursuit--evasion approaches output velocity-level commands and abstract the lower-level quadrotor dynamics \cite{zhangGameDronesMultiUAV2023, xiaoLearningMultipursuitEvasion2024, cetinCounteringDrone3D2022}.
Concurrent work uses body-rate commands with latest-opponent training, reports forgetting and deploys a real pursuer against a virtual evader~\cite{gavinAgileInterception2026}. Its multi-drone follow-up adopts PFSP~\cite{gavinNetCarrying2026}. Our population-based one-versus-one study deploys both learned controllers on real quadrotors. This demonstrates zero-shot transfer of their pursuit--evasion interaction with real flight dynamics on both sides.

\paragraph{Population-based and game-theoretic training}
Self-play and population methods have produced strong results in games and multi-agent RL, including AlphaStar~\cite{vinyals2019grandmaster}, PSRO-style empirical-game training~\cite{lanctot2017unified,mullerGeneralizedTrainingApproach2020}, hide-and-seek~\cite{baker2019emergent}, and competitive drone racing~\cite{pasumarti2025agile}.
AlphaStar's hard PFSP~\cite{vinyals2019grandmaster} weights opponents by $f(w)=(1-w)^p$, where $w$ is the agent's win rate and $p$ controls prioritization. We mix these weights with uniform sampling (see Eq.~\eqref{eq:sampling}).

Pursuit--evasion is asymmetric: pursuers must close and intercept, while evaders receive a survival reward but no bonus when the pursuer leaves the arena. We therefore replace the latest-opponent loop of AMSDRL with frozen quadrotor-controller populations. Our variants sample uniformly (AMSPB-U) or mix hard PFSP with uniform exploration (AMSPB-P). Combined with body-rate RL controllers~\cite{kaufmann2022benchmark}, this lets us study strategy adaptation and transfer both roles to hardware.

\section{METHOD} \label{sec:method}

This section presents the dynamics, game formulation, and PSRO-based training.

\subsection{Quadrotor Dynamics and Control} \label{sec:dynamics}

We model each quadrotor as a rigid body of mass $m$ and diagonal inertia tensor $\mathbf{J} = \mathrm{diag}(J_{xx}, J_{yy}, J_{zz})$.
The state is $\mathbf{x} = (\prescript{W}{}{\mathbf{p}},\, \mathbf{q},\, \prescript{W}{}{\mathbf{v}},\, \boldsymbol{\omega},\, \boldsymbol{\Omega})$, comprising position $\prescript{W}{}{\mathbf{p}} \in \mathbb{R}^3$ in the world frame, attitude quaternion $\mathbf{q} \in S^3$ (body-to-world), world-frame velocity, body-frame angular velocity, and rotor speeds.
The continuous-time dynamics are:
\begin{equation} \label{eq:dynamics}
\begin{bmatrix}
\dot{\mathbf{p}} \\[3pt]
\dot{\mathbf{q}} \\[3pt]
\dot{\mathbf{v}} \\[3pt]
\dot{\boldsymbol{\omega}} \\[3pt]
\dot{\boldsymbol{\Omega}}
\end{bmatrix}
=
\begin{bmatrix}
\mathbf{v} \\[3pt]
\tfrac{1}{2}\,\mathbf{q} \otimes [0,\; \boldsymbol{\omega}^\top]^\top \\[3pt]
\frac{1}{m}\,\mathbf{R}(\mathbf{q})\left(\mathbf{f}_{\text{prop}} + \mathbf{f}_{\text{aero}}\right) + \mathbf{g} \\[3pt]
\mathbf{J}^{-1}\left(\boldsymbol{\tau}_{\text{prop}} - \boldsymbol{\omega} \times \mathbf{J}\boldsymbol{\omega}\right) \\[3pt]
\frac{1}{\tau_{\text{mot}}}\left(\boldsymbol{\Omega}_{\text{cmd}} - \boldsymbol{\Omega}\right)
\end{bmatrix}
\end{equation}
where $\mathbf{R}(\mathbf{q}) \in \mathrm{SO}(3)$ is the rotation matrix from body to world frame, $\mathbf{g}$ is the world-frame gravity vector (Table~\ref{tab:domain_rand}), and $\otimes$ denotes the quaternion product.
Each rotor $i$ produces thrust $f_i = k_\eta \Omega_i^2$ and moment $\tau_i = k_m \Omega_i^2$, where $k_\eta$ and $k_m$ are the thrust and moment coefficients respectively.
The collective propulsive force in the body frame is $\mathbf{f}_{\text{prop}} = [0, 0, \sum_i f_i]^\top$, and the propulsive torque is:
\begin{equation}
    \boldsymbol{\tau}_{\text{prop}} = \mathbf{M}\,[f_1,\, f_2,\, f_3,\, f_4]^\top,
\end{equation}
where $\mathbf{M} \in \mathbb{R}^{3 \times 4}$ is the allocation matrix determined by the X-configuration arm geometry (arm length $l$) and the ratio $k_m / k_\eta$.
The aerodynamic drag in the body frame is $\mathbf{f}_{\text{aero}} = -(\sum_i \Omega_i)\,\mathbf{K}_{\text{aero}}\,\prescript{B}{}{\mathbf{v}}$, with $\mathbf{K}_{\text{aero}} = \mathrm{diag}(k_{xy}, k_{xy}, k_z)$ and $\prescript{B}{}{\mathbf{v}}$ the linear velocity expressed in the body frame.
The motor dynamics are modeled as a first-order system with time constant $\tau_{\text{mot}}$.
In simulation, we use the forward-Euler update
\begin{equation}
    \boldsymbol{\Omega}_{t+1} =
    \alpha_{\text{mot}}\boldsymbol{\Omega}_t
    + (1-\alpha_{\text{mot}})\boldsymbol{\Omega}_{\text{cmd}},\;
    \alpha_{\text{mot}} = 1-\Delta t/\tau_{\text{mot}},
\end{equation}
followed by clamping to the admissible rotor-speed range $[0, \Omega_{\max}]$.
This implicitly limits the maximum thrust to $4\,k_\eta\,\Omega_{\max}^2$.
We use the Crazyflie brushless platform, with parameters coarsely adapted from the Crazyflie brushed identification in~\cite{forster2015system}, and rely on domain randomization during training to handle parameter errors and bridge the sim-to-real gap.

\paragraph{Control architecture}
The RL policies map observations to actions at the policy rate (see Sections~\ref{sec:pursuit_evasion_game} and~\ref{sec:exp_setting} for observations and networks).
Each policy outputs $\mathbf{a}_t \in [-1,1]^4$, interpreted as body-rate and thrust commands:
\begin{equation}
    \omega_{x,y}^{\text{cmd}} = a_{1,2}\cdot \omega_{\max}^{xy}, \quad
    \omega_z^{\text{cmd}} = a_3 \cdot \omega_{\max}^{z}, \quad
    T^{\text{cmd}} = \tfrac{a_4 + 1}{2}\,T_{\max},
\end{equation}
where $\omega_{\max}^{xy}$, $\omega_{\max}^{z}$, and $T_{\max}$ are the roll/pitch rate, yaw rate, and thrust command limits.
This choice follows the benchmark in~\cite{kaufmann2022benchmark}, where body-rate/thrust policies track agile trajectories better than velocity-command policies.
A low-level PID tracks the commanded body rates, converting desired angular acceleration to torque using nominal inertia before rotor allocation.

\subsection{Pursuit--Evasion Game Formulation} \label{sec:pursuit_evasion_game}

We formulate 1v1 pursuit--evasion as a two-player competitive Markov game with opposing terminal objectives between a pursuer ($P$) and an evader ($E$), both governed by~\eqref{eq:dynamics}.
The pursuer aims to intercept, while the evader aims to maximize the time to intercept. Equal vehicle capabilities and a bounded arena let us study adversarial strategy learning without a built-in agility advantage.
An intercept occurs when the Euclidean distance $d_t$ between the agents' geometric centers falls below a capture radius $r_c$:
\begin{equation}
    d_t = \|\prescript{W}{}{\mathbf{p}}_P - \prescript{W}{}{\mathbf{p}}_E\|_2 \leq r_c.
\end{equation}
Each episode ends upon one of four outcomes: pursuer capture, pursuer out-of-bounds, evader out-of-bounds, or evader survival until timeout.
Simultaneously training both agents makes the learning target non-stationary.
Following \cite{xiaoLearningMultipursuitEvasion2024}, we therefore train agents asynchronously: while one agent trains, opponent policies are frozen.

For a fixed distribution over frozen opponent policies, training agent $i \in \{P,E\}$ is a partially observable Markov decision process $G_i=(S,O,A,T,\gamma,R_i)$.
Here $S$ contains both vehicle states, the sampled opponent identity, controller and task states, randomized parameters, and histories needed for delays and rewards. $O$ is the observation space, $A=[-1,1]^4$ the action space, and $T$ the transition under~\eqref{eq:dynamics} and the frozen opponent. The discount factor is $\gamma$ and the reward is $R_i$.
The objective to maximize is
\begin{equation}
    J_i(\pi_i;\mu_{-i}) =
    \mathbb{E}_{\pi_i,\,\pi_{-i}\sim\mu_{-i}}
    \left[\sum_{t=0}^{\infty}\gamma^t R_i(s_t,a_t)\right],
\end{equation}
where $\mu_{-i}$ is the sampled opponent distribution.
Both agents use the same observation template, but all body-frame and relative quantities are expressed from the observing agent's frame:
\begin{equation} \label{eq:obs}
    \mathbf{o}_t = [\mathbf{R}_{\text{flat}},\; \prescript{B}{}{\mathbf{v}},\; \boldsymbol{\omega},\; \prescript{W}{}{\mathbf{p}},\; \prescript{B}{}{\Delta\mathbf{p}},\; \prescript{B}{}{\Delta\mathbf{v}}] \in \mathbb{R}^{24}.
\end{equation}
For observing agent $i$ and opponent $j$, $\Delta\mathbf{p}=\prescript{W}{}{\mathbf{p}}_j-\prescript{W}{}{\mathbf{p}}_i$ and $\Delta\mathbf{v}=\prescript{W}{}{\mathbf{v}}_j-\prescript{W}{}{\mathbf{v}}_i$, transformed to the body frame through $\mathbf{R}(\mathbf{q}_i)^\top$.
The terms in~\eqref{eq:obs} are the flattened rotation matrix, body-frame velocity and angular velocity, world-frame position, and opponent-relative position and velocity.

\paragraph{Rewards}\label{sec:rewards}
The rewards use task outcomes with limited shaping and regularization.
Both agents share
\begin{equation}
    \begin{aligned}
    r^{\text{reg}}_t={}&-\kappa_{\text{br}}(\|\boldsymbol{\omega}_{xy}\|_2+|\omega_z|)\\
    &-\kappa_{\text{sm}}\|\mathbf{a}^{\omega}_t-\mathbf{a}^{\omega}_{t-1}\|_2
    -\kappa_{\text{sm}}|a_{4,t}-a_{4,t-1}|-\kappa_b b_t,
    \end{aligned}
\end{equation}
where $\mathbf{a}^{\omega}_t$ denotes the three body-rate action components, $a_{4,t}$ is the thrust action, $\boldsymbol{\omega}_{xy}$ contains roll/pitch rates, $\kappa_{\text{br}}$ penalizes high body rates, $\kappa_{\text{sm}}$ penalizes rapid action changes, and $b_t$ is an out-of-bounds indicator for the learning agent with penalty weight $\kappa_b$.

For the pursuer, the reward is
\begin{equation} \label{eq:reward_p}
    r^P_t = \kappa_c \cdot \delta_t \cdot \beta_t
    + \kappa_{\text{app}}(\bar d_{t-1}-d_t)
    + r^{\text{reg}}_t,
\end{equation}
where $\delta_t = \mathbf{1}_{\{d_t \leq r_c\}}$ indicates capture.
The factor $\beta_t$ reduces the capture bonus for unsafe high-closing-speed interceptions:
\begin{equation}\label{eq:closing_reward}
    \beta_t = \exp\!\left(-\alpha_{\text{close}} \cdot \max(0,\; v_{\text{close}} - v_{\text{safe}})\right).
\end{equation}
Here $v_{\text{close}} = \max(0,-\hat{\mathbf{u}}^\top (\prescript{W}{}{\mathbf{v}}_E - \prescript{W}{}{\mathbf{v}}_P))$ is the closing speed along the unit line of sight $\hat{\mathbf{u}}=(\mathbf{p}_E-\mathbf{p}_P)/d_t$, $v_{\text{safe}}$ is the closing-speed threshold, and $\alpha_{\text{close}}$ controls how quickly the capture bonus decays above that threshold.
This gives the full capture bonus only when the closing speed is moderate, discouraging frontal high-impact interceptions.
The approach term rewards reductions from smoothed distance, $\bar d_t=\alpha_d\bar d_{t-1}+(1-\alpha_d)d_t$, with factor $\alpha_d$ and $\bar d_0=d_0$.

For the evader, the reward is
\begin{equation} \label{eq:reward_e}
    r^E_t = \kappa_s - \kappa_c \cdot \delta_t + r^{\text{reg}}_t,
\end{equation}
where $\kappa_s$ is a per-timestep survival bonus and $\kappa_c$ is the capture reward scale.
The evader reward avoids distance-based shaping, letting evasion emerge from capture and survival.
Since arena limits are deployment constraints, each agent is penalized only for its own violation, with no reward for opponent boundary exits.

\paragraph{Termination and truncation}
Capture and the learning agent's own boundary violation are terminal, setting the PPO bootstrap value to zero. Opponent boundary exits and timeouts also end the episode but are treated as truncations, using the critic estimate $V_\phi(\mathbf{o}_{t+1})$ for the next observation.
For either role, this preserves an estimate of the learning agent's remaining return when its opponent leaves the arena.

\subsection{Population-Based Training via PSRO} \label{sec:psro}

\begin{figure*}[t]
    \vspace{0.04in}
    \centering
    \def\svgwidth{\textwidth}
    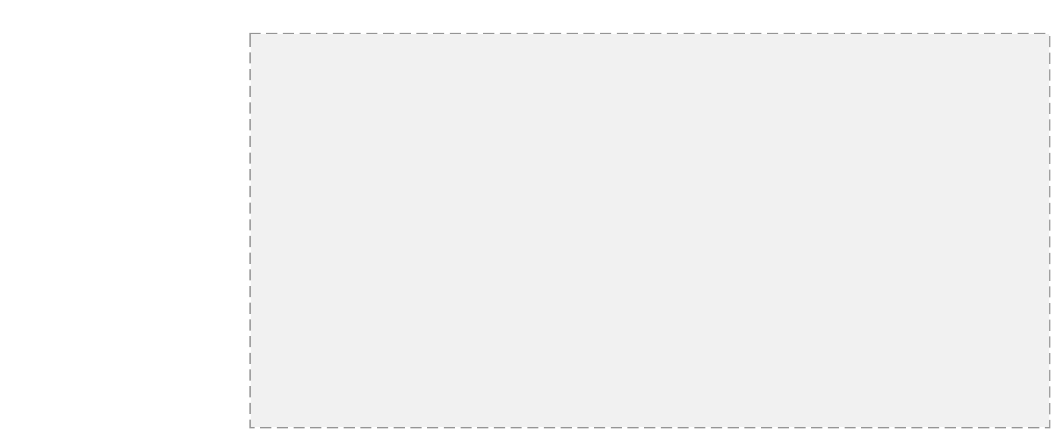
    \caption{AMSPB-P training: (a) pretrain against trajectory evaders and FRPN-V, (b) evaluate the previous policy, (c) sample opponents using~\eqref{eq:sampling}, and (d) train a PPO response and append it to the population. Baseline controllers remain available throughout.}
    \label{fig:training_pipeline}
\end{figure*}

Training against a single opponent is prone to forgetting and cycling.
AMSPB-P addresses this by adapting Policy-Space Response Oracles (PSRO)~\cite{lanctot2017unified,mullerGeneralizedTrainingApproach2020} from empirical-game strategy learning to continuous quadrotor control.
In PSRO, new oracles are approximate best responses to mixtures over previously discovered strategies.
In our setting, each strategy is a neural body-rate quadrotor controller trained with PPO under the dynamics and randomization above.
Let $\pi_P^{(k)}$ and $\pi_E^{(k)}$ denote the pursuer and evader policies produced at generation $k$, and let $\Pi_P^{(k)}$ and $\Pi_E^{(k)}$ denote the corresponding populations after generation $k$.
Rather than constructing a full empirical payoff table and running a meta-solver at every generation, AMSPB-P builds the mixture directly from rollout-estimated win rates using hard PFSP~\cite{vinyals2019grandmaster} with uniform exploration.

\paragraph{Training procedure}
Training proceeds over $K$ generations (see Fig.~\ref{fig:training_pipeline}).
At generation 0, each role $i$ has population $\Pi_i^{(0)}=H_i\cup\{\pi_i^{(0)}\}$, where $H_i$ contains trajectory-based or reactive controllers and $\pi_i^{(0)}$ is pretrained against role-appropriate opponents (Section~\ref{sec:baselines}).
At each subsequent generation $k$, both roles train independently against previous-generation opponent populations: the evader trains against $\Pi_P^{(k-1)}$ to obtain $\pi_E^{(k)}$, and the pursuer trains against $\Pi_E^{(k-1)}$ to obtain $\pi_P^{(k)}$.
Thus, neither oracle observes a policy from the same generation while it is being trained.
Both populations are then updated: $\Pi_E^{(k)} = \Pi_E^{(k-1)} \cup \{\pi_E^{(k)}\}$ and $\Pi_P^{(k)} = \Pi_P^{(k-1)} \cup \{\pi_P^{(k)}\}$.
This differs from AMSDRL~\cite{xiaoLearningMultipursuitEvasion2024}, where each agent trains against only the latest opponent, including the current generation's counterpart.

\paragraph{Prioritized opponent sampling}
Before training agent $i$ at generation $k$, we estimate the win rate of the current policy $\pi_i^{(k-1)}$ against each frozen opponent $\pi_j \in \Pi_{-i}^{(k-1)}$ using rollout evaluations under randomized training conditions.
The challenge score of opponent $j$ is $c_j = 1 - w_j$, where $w_j$ is the measured win rate.
We use the hard-PFSP priority $c_j^{\eta_s}$ from~\cite{vinyals2019grandmaster}, normalize the weights, and mix them with uniform exploration:
\begin{equation} \label{eq:sampling}
    p_j = \frac{\gamma_s}{N}
    + (1-\gamma_s)\,
    \frac{c_j^{\eta_s}}{\sum_{\ell=1}^{N} c_\ell^{\eta_s}},
\end{equation}
where $N$ counts opponent sampling entries, with the trajectory controllers pooled into one entry. Each entry has a shared score, $\eta_s$ controls prioritization, and $\gamma_s$ ensures uniform coverage.
The subscript $s$ distinguishes sampler parameters from the discount parameter $\gamma$. If all challenge scores are zero, the normalized challenge term is replaced by a uniform distribution, and at $\eta_s=0$ all priority weights are one.
The pursuer win rate counts capture or evader out-of-bounds, whereas the evader win rate counts timeout or pursuer out-of-bounds.
The sampled opponent identity is not included in the observation, so the oracle is trained as a single policy against the mixture rather than as a separate policy conditioned on opponent identity.
AMSPB-U (U for uniform) samples the same population uniformly, recovered by $\eta_s=0$ or $\gamma_s=1$. With $\gamma_s=0$ the rule is pure hard PFSP. For $\eta_s>0$ and $\gamma_s<1$, it concentrates training on challenging opponents while maintaining the uniform component.

\paragraph{Policy training}
The actor and critic are separate feed-forward multi-layer perceptrons (MLPs) (see Section~\ref{sec:exp_setting}).
The actor outputs a Gaussian distribution over actions; the critic is deterministic.
We use symmetric actor-critic training, so both actor and critic receive $\mathbf{o}_t$.
Training uses the standard clipped PPO objective~\cite{schulmanProximalPolicyOptimization2017}.
We also use a stage-specific Kullback--Leibler (KL) threshold to adapt the learning rate during training, together with entropy regularization to maintain exploration as the opponent distribution changes.

Each generation warm-starts from the full previous agent, including optimizer and running-standardization state.

\section{EXPERIMENTS} \label{sec:experiments}

We first evaluate AMSPB-P against AMSDRL~\cite{xiaoLearningMultipursuitEvasion2024} and heuristic baselines in simulation, and then validate the trained policies on real Crazyflie quadrotors.

\subsection{Experimental Settings} \label{sec:exp_setting}

\paragraph{Simulation}
We use Isaac Lab~\cite{mittal2025isaac} as the training framework, Isaac Sim for GPU-accelerated physics and SKRL for PPO~\cite{serrano2023skrl}.
The simulation runs at 500~Hz and the RL policy at 50~Hz.
Each oracle is trained for 50M frames on 6144 parallel environments, with another $50$M per role for shared pretraining.
Main simulation parameters and domain randomization ranges are reported in Table~\ref{tab:domain_rand}.
Domain randomization is applied throughout training.
Agents spawn with random orientation ($\pm 5^\circ$ roll/pitch, $\pm 180^\circ$ yaw) and initial velocity ($\pm 0.1$~m/s), at a minimum inter-agent separation of $0.5$~m and a minimum clearance of $0.5$~m from the arena boundaries.
The arena's vertical bounds are $z\in[0.1,2.2]$~m.

\begin{table}[!t]
    \vspace{0.04in}
    \centering
    \footnotesize
    \setlength{\tabcolsep}{3.5pt}
    \renewcommand{\arraystretch}{0.95}
    \caption{Main simulation settings and domain randomization ranges (multipliers of nominal values). Rate-gain vectors are ordered roll, pitch, yaw.}
    \begin{tabular}{@{}lll@{}}
    \toprule
    Parameter & Nominal value & Multiplier \\
    \midrule
    Arena & $3.6{\times}4.8{\times}2.1$~m & fixed \\
    Gravity $\mathbf{g}$ & $(0,0,-9.81)$~m/s$^2$ & fixed \\
    Capture radius $r_c$ & $0.1$~m & fixed \\
    Horizon; policy/rate ctrl. & $10$~s; $50/500$~Hz & fixed \\
    Action delay & $20$~ms (one policy step) & fixed \\
    Asset mass $m$ & $39.3$~g & $[0.9,1.1]$ \\
    Body inertia $\mathrm{diag}(\mathbf{J})$ & $(3,3,4)10^{-5}$~kg\,m$^2$ & $[0.8,1.2]$ \\
    Thrust $k_\eta$ & $1.2\!\times\!10^{-8}$~N\,s$^2$/rad$^2$ & $[0.8,1.2]$ \\
    Moment $k_m$ & $2.4\!\times\!10^{-10}$~N\,m\,s$^2$/rad$^2$ & $[0.8,1.2]$ \\
    Aero.\ drag $k_{xy}$ & $9.1785\!\times\!10^{-7}$~N\,s$^2$/(m\,rad) & $[0.5,2.0]$ \\
    Aero.\ drag $k_z$ & $1.0311\!\times\!10^{-6}$~N\,s$^2$/(m\,rad) & $[0.5,2.0]$ \\
    Motor $\tau_{\text{mot}}$ & $5$~ms & $[0.8,1.2]$ \\
    Rate PID $K_p$ & $(200,200,120)$~s$^{-1}$ & $[0.9,1.1]$ \\
    Rate PID $K_i$ & $(400,400,16.7)$~s$^{-2}$ & $[0.9,1.1]$ \\
    Rate PID $K_d$ & $(2.5,2.5,0)$ & $[0.9,1.1]$ \\
    Rate integral clamp & $\pm(33.3,33.3,166.7)^\circ$ & fixed \\
    Arm length $l$; X angle & $0.05$~m; $45^\circ$ & fixed \\
    Rotor limit $\Omega_{\max}$ & $4200$~rad/s & fixed \\
    Rate limits $\omega_{\max}^{xy},\omega_{\max}^{z}$ & $(100,200)^\circ$/s & fixed \\
    Command limit $T_{\max}$ & $1.2$~N & fixed \\
    Rotor-limited force & $4k_\eta\Omega_{\max}^2=0.847$~N & via $k_\eta$ \\
    \bottomrule
    \end{tabular}
    \label{tab:domain_rand}
\end{table}

\paragraph{Population training}
We run $K{=}13$ generations using $\gamma_s = 0.7$ and $\eta_s = 2$ for nominal AMSPB-P.
Pretraining uses trajectory evaders and an FRPN-V pursuer (Section~\ref{sec:baselines}).
Before each generation, we estimate opponent win rates under randomized training conditions using $6144$ parallel evaluation environments shared across the current population.
The main comparison uses three training runs per method, each with $1.30$B adversarial-training transitions and $100$M shared pretraining transitions. Sampler evaluation adds $76.8$--$78.5$M transitions ($5.9$--$6.0\%$ of the adversarial budget) to AMSPB-P without updating the policies.

\paragraph{Policy and reward parameters}
All learned policies use separate actor and critic MLPs with layers $[512,256,256,128]$ and exponential linear unit (ELU) activations.
PPO uses a discount-factor $\gamma{=}0.998$, learning rate $10^{-4}$ with a KL-adaptive schedule (thresholds $0.008$ for pretraining and $0.012$ for adversarial training), $16$ rollout steps, $8$ epochs, $16$ mini-batches, generalized advantage estimation (GAE) with $\lambda{=}0.95$, clip ratio $0.2$, entropy coefficients $0.01$/$0.001$ for pursuer/evader, and value loss scale $1.0$.
The reward coefficients are $\kappa_{\text{br}}{=}0.001$, $\kappa_{\text{sm}}{=}0.002$, $\kappa_b{=}10.0$, $\kappa_c{=}10.0$, $\kappa_{\text{app}}{=}0.25$, and $\kappa_s{=}0.025$.
The closing-speed capture penalty uses $v_{\text{safe}}{=}1.5$~m/s and $\alpha_{\text{close}}{=}3.0$, with distance smoothing $\alpha_d=0.8$.

For final cross-method evaluation (Fig.~\ref{fig:cross_method_bars}), we run 1024 episodes per matchup and seed.
The initial positions are $(0, -1.5, 1)$~m for the pursuer and $(0, 1.5, 1)$~m for the evader, randomized within a $0.1$~m sphere.
No domain randomization is applied during evaluation and actors are set deterministic.

\subsection{Baselines} \label{sec:baselines}

We compare against both heuristic and learning-based baselines.
Velocity-based baselines use a $100$~Hz velocity loop to set attitude and thrust, followed by attitude control and the same $500$~Hz body-rate loop (Section~\ref{sec:dynamics}). Learned controllers instead command body rates and thrust directly.

\paragraph{FRPN-inspired velocity pursuer (FRPN-V)}
Using world-frame relative states, we adapt~\cite{pliskaSafeMidAirDrone2024} as
\begin{equation}
\mathbf{v}_{\mathrm{cmd}}=G\left((1-W)
\frac{\Delta\mathbf{p}+\Delta\mathbf{v}t_{\mathrm{go}}}{t_{\mathrm{go}}^2+\epsilon_t}
+W\Delta\mathbf{p}\right),
\end{equation}
Here $t_{\mathrm{go}}=\|\Delta\mathbf{p}\|/(\|\Delta\mathbf{v}\|+\epsilon_v)$, with numerical stabilizers $\epsilon_v=10^{-6}$~m/s and $\epsilon_t=10^{-6}$~s$^2$.
We set $G=4$, $W=0.25$, tuned against trajectory and reactive APF evaders, not learned opponents, and clip each velocity component to $\pm(1.5,1.5,1.0)$~m/s.
This variant supplies evader pretraining, but its velocity commands and tuning differ from acceleration-command FRPN, so our results concern this adaptation.

\paragraph{Trajectory-based evaders}
These non-reactive evaders follow hovering, circular, or lemniscate (figure-eight) trajectories.
Circles and lemniscates sample radii $[0.93,1.55]$ and $[0.725,1.45]$~m, respectively, and angular parameters $[1/r_{\max},2/r_{\max}]$~rad/s, where $r_{\max}$ is the family maximum.
The paths have randomized phase, orientation, height parameter $[0.85,1.65]$~m and elliptical deformation. Hover positions are randomized within the same arena.
These paths use bounds inset by $0.25$~m and enable pursuers to anticipate curvilinear motion.

\paragraph{Artificial Potential Field (APF) evader}
This reactive evader uses repulsive fields from the pursuer and arena boundaries~\cite{maarifArtificialPotentialField2021}.
The pursuer repulsion is $\mathbf{f}_P = k_P \mathbf{r} / \|\mathbf{r}\|^{\alpha_P}$ inside its activation radius, with analogous wall terms. The normalized total force is tracked as a velocity reference, where $\mathbf{r}={}^W\mathbf{p}_E-{}^W\mathbf{p}_P$ points away from the pursuer.
We use $k_P{=}0.5$, $k_W{=}0.5$, $\alpha_P{=}2$, and $\alpha_W{=}2$, kept fixed for all evaluations. Activation distances are $4$~m and speed is $1.5$~m/s, with the same inset bounds. APF is excluded from policy-training opponents but used in sampler selection.

\paragraph{AMSDRL}
We re-implement \cite{xiaoLearningMultipursuitEvasion2024} in our environment with body-rate control for fair comparison.
AMSDRL trains each agent against only the latest opponent, including the current generation's counterpart, corresponding to the single-opponent asynchronous setting.

\paragraph{AMSPB-U}
Our uniform variant, AMSPB-U, samples the same population equally.

\subsection{Results} \label{sec:results}

\begin{figure}[!t]
    \centering
    \includegraphics[width=\linewidth]{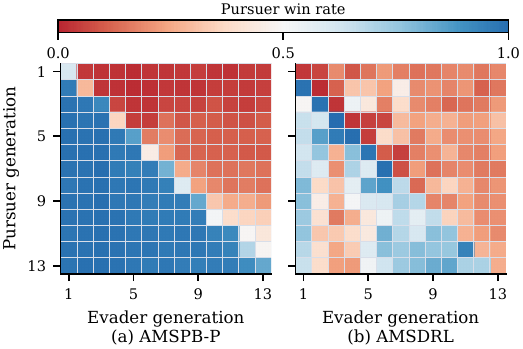}
    \caption{Pursuer win-rate heatmaps across 13 generations averaged across the 3 training seeds. Cell $(i,j)$ is the win rate of pursuer generation $i$ against evader generation $j$.}
    \label{fig:heatmaps}
\end{figure}

\begin{figure}[!t]
    \centering
    \includegraphics[width=0.96\linewidth]{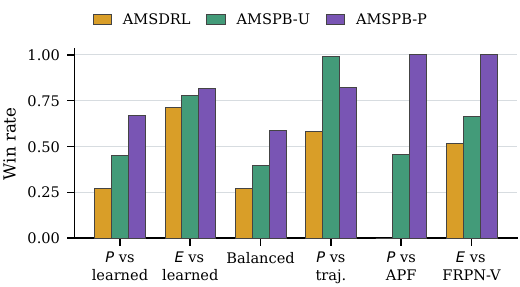}
    \caption{
    Generation-13 win rates averaged over three seeds. Learned scores face the other two methods' final opposite-role policies, and Balanced is the per-seed minimum of the two role scores. Remaining groups use trajectory and APF evaders, and the FRPN-V pursuer.
    }
    \label{fig:cross_method_bars}
\end{figure}

\begin{figure}[!t]
\centering
\includegraphics[width=\columnwidth]{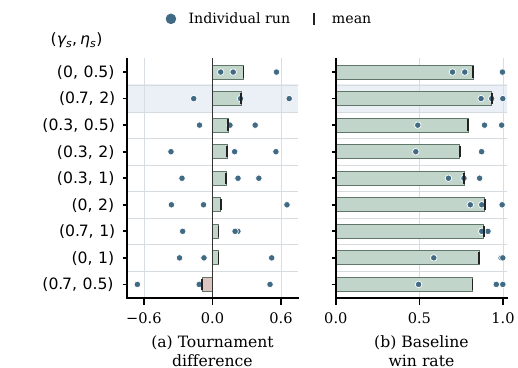}
\caption{Sampler ablation ($n=3$ runs per setting). (a) Role-summed tournament-score difference relative to AMSPB-U. Generation-13 policies face AMSDRL and the compared method's generations $1,5,9,13$ ($128$ episodes per matchup), so the reference varies by setting. (b) Role-balanced win rate against baseline controllers. Dots show individual runs, ticks show means, and shading marks the deployed setting.}
\label{fig:amspbh_ablation}
\end{figure}

\begin{figure*}[!t]
    \vspace{0.04in}
    \centering
    \includegraphics[width=0.78\textwidth]{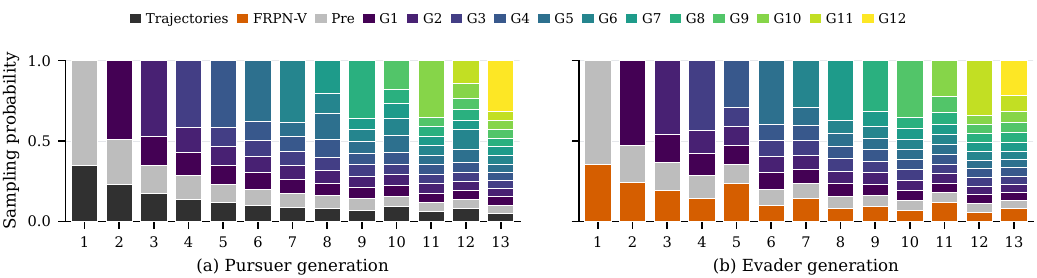}
    \caption{
    AMSPB-P opponent sampling distributions, seed $42$. Pursuer training (a) starts against fixed trajectories, while evader training (b) starts against FRPN-V. The prioritized sampler emphasizes difficult newer opponents while keeping older ones active. \emph{Pre} denotes the pretraining policy.
    }
    \label{fig:opponent_mixture}
\end{figure*}

\begin{figure*}[!t]
\centering
\includegraphics[width=\textwidth,trim=0 70 0 80,clip]{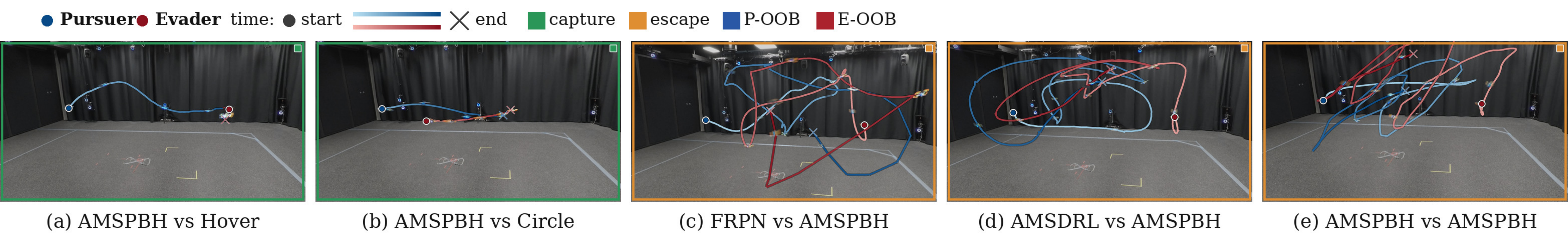}
\par\smallskip
{\footnotesize\makebox[.2\textwidth]{(a) AMSPB-P vs Hover}%
\makebox[.2\textwidth]{(b) AMSPB-P vs Circle}%
\makebox[.2\textwidth]{(c) FRPN-V vs AMSPB-P}%
\makebox[.2\textwidth]{(d) AMSDRL vs AMSPB-P}%
\makebox[.2\textwidth]{(e) AMSPB-P vs AMSPB-P}}
\caption{Representative hardware trajectories. Blue/red curves show pursuer/evader image-plane motion, with lighter colors denoting earlier positions and darker colors denoting later positions. Filled dots mark initial positions and crosses mark final positions of the displayed interval. Panel border colors indicate capture (green) or no capture (orange) within that interval. Panel (e) ends at $10$~s before the capture reported in Table~\ref{tab:real_world_matchups}.}
\label{fig:experimental_trajectories}
\end{figure*}

We report simulation results in Figs.~\ref{fig:heatmaps}--\ref{fig:opponent_mixture} and hardware results in Fig.~\ref{fig:experimental_trajectories} and Table~\ref{tab:real_world_matchups}.
\paragraph{AMSPB-P vs.\ AMSDRL: population-based training}
Fig.~\ref{fig:heatmaps} compares pursuer win-rate matrices for AMSPB-P and AMSDRL across 13 generations.
AMSPB-P shows a clear triangular pattern: later pursuers retain competence against older evaders while adapting to newer ones.
AMSDRL has a weaker pattern, consistent with forgetting and cycling observed in adversarial learning~\cite{vinyals2019grandmaster,gavinAgileInterception2026}.

\paragraph{Cross-evaluation against diverse opponents}
Fig.~\ref{fig:cross_method_bars} compares the final-generation policies against learned, trajectory-based and reactive opponents.
The Balanced score is the per-seed minimum of the pursuer and evader learned-opponent win rates. AMSPB-P achieves the highest mean Balanced score ($0.59$, versus $0.40$ for AMSPB-U and $0.27$ for AMSDRL).
AMSPB-U is higher against trajectory evaders, while AMSPB-P performs best against APF. The trajectory gap mainly reflects lower AMSPB-P capture of the hovering evader. Hover, circle and figure-eight share one sampling score, so success on the moving paths can reduce hover exposure. Three diagnostic AMSPB-P split runs give hover its own sampling entry after unchanged pretraining, restoring hover capture. Main results retain the pooled sampler.
AMSDRL is weaker in most groups, highlighting the limited generalization of single-opponent training.
These results suggest that exposure to a mixture of learned and heuristic strategies improves performance against opponents outside the policy-training loop.

\paragraph{Sampler ablation and auto-curriculum}
Fig.~\ref{fig:amspbh_ablation} compares prioritized sampling with our uniform population variant, AMSPB-U.
Most tested configurations match or exceed uniform sampling on average, with the benefit depending on the setting and opponents.
The remaining experiments use $\gamma_s=0.7$, $\eta_s=2$, retaining $70\%$ uniform sampling while directing $30\%$ toward squared opponent difficulty.
This combines continued exposure to the whole population with more frequent training against difficult opponents.

Fig.~\ref{fig:opponent_mixture} illustrates the opponent sampling distribution from~\eqref{eq:sampling}.
As training progresses, probability shifts toward recent challenging opponents while retaining non-zero mass on earlier and heuristic opponents.
This creates an auto-curriculum while keeping earlier policies and baseline controllers available.

\paragraph{Broader engagement conditions}
Across $60$ starting geometries near the arena center, with varied separation, direction and relative height, mean Balanced scores remain ordered AMSPB-P, AMSPB-U, AMSDRL ($0.44$, $0.19$, $0.15$).
Repeating these geometries across the arena gives broadly similar mean win rates across separations and heights within each learned matchup.
With starts distributed throughout the nominal arena, extending the same rollouts from $10$ to $20$~s raises AMSPB-P capture from $0.43$ to $0.44$ against the equally weighted three learned families, including its own, and from $0.71$ to $0.74$ against APF. It remains highest in both comparisons.

We also retrain all methods at $1.5$ times each arena dimension with the same three seeds and budget.
Comparing small-arena/$10$~s and large-arena/$15$~s evaluations on correspondingly distributed starts, AMSPB-P maintains APF capture at $0.71$ and improves FRPN-V evasion from $0.68$ to $0.86$.
Both population methods exceed AMSDRL in mean role win rate against the three learned families.

\paragraph{Real-hardware deployment}
We deploy the policies on two Crazyflie brushless quadrotors without hardware fine-tuning. Both take off with the cascaded PID controller and hover at the benchmark initial positions.
A motion-capture system provides position and orientation at 100~Hz.
After switching to the learned policy or baseline, the offboard controller sends commands at 50~Hz for a 12~s engagement.
\finalchange{Hardware capture uses a $0.15$~m radius instead of the $0.1$~m simulation radius in Table~\ref{tab:domain_rand}: simulation disables collisions between the vehicles, whereas on hardware the propeller guards can touch before the centers come within $0.1$~m on horizontal approaches.}
All trials use full batteries to reduce battery-dependent artifacts, and no policy weights, reward parameters, or controller gains are tuned on hardware.

\begin{table*}[!t]
\vspace{0.04in}
\centering
\footnotesize
\setlength{\tabcolsep}{4.2pt}
\renewcommand{\arraystretch}{1.0}
\caption{Real-world pursuit--evasion matchup summary.}
\label{tab:real_world_matchups}
\begin{tabular}{@{}ll@{\hspace{6pt}}rrrr@{\hspace{7pt}}rr@{\hspace{7pt}}rr@{\hspace{7pt}}rr@{\hspace{3pt}}rr@{\hspace{7pt}}rr@{\hspace{3pt}}rr@{}}
\toprule
& & \multicolumn{4}{c}{Outcome} & \multicolumn{2}{c}{$T$ [s]} & \multicolumn{2}{c}{$v_{\text{close}}\mid C$ [m/s]} & \multicolumn{4}{c}{$v$ [m/s]} & \multicolumn{4}{c}{$\omega$ [rad/s]} \\
\cmidrule(lr){3-6}\cmidrule(lr){7-8}\cmidrule(lr){9-10}\cmidrule(lr){11-14}\cmidrule(l){15-18}
Pursuer & Evader & C & E & P-OOB & E-OOB & avg & max & avg & max & \multicolumn{2}{c}{$P$} & \multicolumn{2}{c}{$E$} & \multicolumn{2}{c}{$P$} & \multicolumn{2}{c}{$E$} \\
 & & & & & & & & & & avg & max & avg & max & avg & max & avg & max \\
\midrule
\textbf{AMSPB-P} & Hover & \textbf{5} & 0 & 0 & 0 & 2.48 & 2.93 & 1.26 & 2.32 & 1.42 & 2.64 & 0.07 & 0.16 & 3.06 & 5.02 & 0.16 & 3.64 \\
 & Circle & \textbf{5} & 0 & 0 & 0 & 1.93 & 2.21 & 0.90 & 1.69 & 1.67 & 2.49 & 0.90 & 1.19 & 3.25 & 5.82 & 0.55 & 3.06 \\
 & APF \cite{maarifArtificialPotentialField2021} & \textbf{4} & 1 & 0 & 0 & 5.74 & 11.99 & 0.86 & 2.60 & 1.44 & 2.47 & 0.96 & 1.58 & 2.95 & 5.96 & 1.10 & 3.09 \\
 & Manual & \textbf{3} & 0 & 0 & 0 & 3.05 & 3.15 & 1.01 & 2.56 & 1.62 & 2.87 & 0.98 & 1.76 & 3.12 & 4.76 & 1.66 & 4.49 \\
 & \textbf{AMSPB-P} & \textbf{5} & 0 & 0 & 0 & 4.32 & 10.92 & 0.94 & 2.42 & 1.94 & 3.13 & 1.77 & 2.95 & 3.39 & 5.78 & 2.68 & 4.26 \\
 & \textbf{AMSPB-U} & \textbf{2} & 0 & 2 & \textbf{1} & 5.77 & 6.04 & 0.82 & 2.48 & 2.10 & 3.71 & 1.67 & 3.70 & 3.24 & 5.52 & 1.98 & 4.46 \\
 & AMSDRL \cite{xiaoLearningMultipursuitEvasion2024} & \textbf{3} & 0 & 0 & 0 & 6.17 & 11.98 & 0.76 & 2.57 & 1.98 & \textbf{4.87} & 1.55 & 3.38 & 2.78 & 5.01 & 2.81 & 5.20 \\
\midrule
\textbf{AMSPB-U} & \textbf{AMSPB-P} & 1 & \textbf{2} & 0 & 0 & 11.49 & 11.99 & 0.58 & 2.65 & 1.52 & 3.19 & 1.40 & 3.27 & 3.08 & \textbf{6.08} & 1.84 & 4.30 \\
\midrule
Manual & \textbf{AMSPB-P} & 0 & \textbf{3} & 0 & 0 & 12.00 & 12.00 & -- & -- & 1.02 & 1.64 & 1.49 & 3.65 & 1.11 & 3.91 & 1.96 & 4.91 \\
AMSDRL \cite{xiaoLearningMultipursuitEvasion2024} & \textbf{AMSPB-P} & 2 & \textbf{1} & 0 & 0 & 8.94 & 11.99 & 0.73 & 3.06 & 1.79 & 3.49 & 1.58 & 3.29 & 2.80 & 5.12 & 2.12 & 3.98 \\
FRPN-V & \textbf{AMSPB-P} & 0 & \textbf{2} & 0 & 0 & 12.00 & 12.00 & -- & -- & 1.27 & 1.99 & 1.51 & 3.28 & 1.34 & 3.52 & 2.07 & 4.89 \\
\bottomrule
\end{tabular}
\par\vspace{0.7ex}
\noindent\begin{minipage}{\textwidth}
\footnotesize C: capture, E: timeout escape, P-OOB/E-OOB: pursuer/evader out-of-bounds, respectively. Speeds are reported separately for pursuer ($P$) and evader ($E$). $T$ is duration. Avg averages trace means, and max is the largest logged value. $v_{\text{close}}\mid C$ is over capture trials only, throughout traces rather than at capture (Fig.~\ref{fig:paper_hardware_envelope}). AMSPB-P versus Circle, Manual and AMSPB-U have $4$, $2$ and $4$ traces, respectively, with one per outcome elsewhere. Bold outcome counts (left part of table) mark AMSPB-P wins and bold speeds mark table maxima.
\end{minipage}
\end{table*}

Fig.~\ref{fig:experimental_trajectories} and Table~\ref{tab:real_world_matchups} summarize the physical experiments.
The manual trials use high-level velocity commands generated by a human, providing motion different from the training opponents. They are included to provide qualitative checks of the policies' reactions, not a matched-agility comparison with the learned body-rate controllers.
Across five AMSPB-P flights, pursuit aligns better with the evader's $0.3$~s constant-velocity extrapolation than its current position. Hiding the evader's velocity from the pursuer slows pursuit and reduces captures in paired simulations, supporting motion-dependent interception without establishing explicit prediction.

\begin{figure}[!t]
    \centering
    \includegraphics[width=0.82\columnwidth]{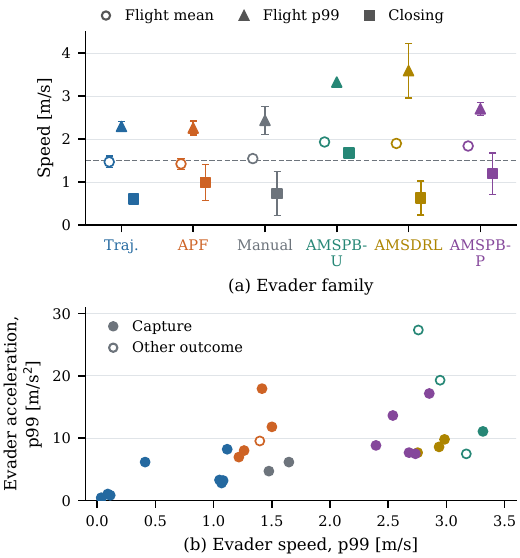}
    \caption{AMSPB-P hardware measurements. (a) Family means of per-flight mean/p99 pursuit speed and capture closing speed ($24$ captures), with sample-SD whiskers. Dashed: $v_{\mathrm{safe}}=1.5$~m/s. (b) Evader p99 speed/acceleration in $28$ logged flights, filled/open for capture/other outcomes. Colors identify evader families, with Traj. denoting hover and circle. Peaks are within-flight $99$th percentiles from smoothed positions.}
    \label{fig:paper_hardware_envelope}
\end{figure}

Across $42$ flights, the AMSPB-P pursuer reaches a logged peak speed of $4.87$~m/s.
Fig.~\ref{fig:paper_hardware_envelope} separates flight speed from closing speed at capture.

\Needspace{4\baselineskip}
Among $24$ logged AMSPB-P captures, evaders reach robust peak speeds up to $3.31$~m/s and accelerations up to $17.95$~m/s$^2$.
Closing speed at capture has median $0.73$~m/s and maximum $1.68$~m/s, consistent with the moderate approaches encouraged by~\eqref{eq:closing_reward}.
The cited learning-based flight studies use $1$~m/s limits~\cite{xiaoLearningMultipursuitEvasion2024,chenOnlinePlanningMultiUAV2025}, whereas onboard perception and physical net capture have enabled interception of a $5$~m/s target with up to $4.9$~m/s$^2$ acceleration~\cite{pliskaSafeMidAirDrone2024}.

\section{CONCLUSIONS AND FUTURE WORK} \label{sec:conclusions}

We introduced uniform and prioritized population-based training for 1v1 quadrotor pursuit--evasion with body-rate control.
By training approximate best responses against hard-PFSP mixtures with uniform exploration, AMSPB-P empirically reduces catastrophic forgetting and strategy cycling in our evaluations.
Compared to single-opponent training (AMSDRL), it retains competence against older strategies while adapting to new ones.
We validated the trained policies on real Crazyflie brushless quadrotors across $42$ flights, demonstrating zero-shot sim-to-real transfer with both learned roles deployed on physical vehicles.

The frozen population is both a training memory and a curriculum, exposing each controller to old, recent, heuristic and learned strategies.
Two limitations remain.
First, the feed-forward MLP policies limit memory and capacity in the partially observable mixture setting. Recurrent or history-conditioned architectures may improve robustness.
Second, future work will incorporate onboard perception and obstacle-aware observations to extend beyond the current motion-capture-localized, obstacle-free setting.


\bibliographystyle{IEEEtran}
\bibliography{main,Learned_Reactive_Controller} 

\end{document}